\documentclass{article}

\usepackage[table]{xcolor}
\usepackage{microtype}
\usepackage{graphicx}
\usepackage{subfigure}
\usepackage{booktabs} % for professional tables

\usepackage{hyperref}
 
\usepackage[preprint]{icml2025}

\usepackage{amsmath}
\usepackage{amssymb}
\usepackage{mathtools}
\usepackage{amsthm}
\usepackage{xspace} % Ensure xspace is included

\usepackage[capitalize,noabbrev]{cleveref}

\theoremstyle{plain}

\theoremstyle{definition}

\theoremstyle{remark}

\usepackage[disable,textsize=tiny]{todonotes}
\usepackage{multirow}
\usepackage{cancel}
\newcommand\Ccancel[2][black]{\renewcommand\CancelColor{\color{#1}}\cancel{#2}}

\newcommand{\solidline}[2][black]{\textcolor{#1}{\rule[0.5ex]{#2}{1.5pt}}}
\newcommand{\dottedline}[2][black]{ \textcolor{#1}{\rule[0.5ex]{#2}{1pt} \rule[0.5ex]{#2}{1pt}}}

\newcommand{\redsquare}[1]{%
    \begin{tikzpicture}
        \fill[red] (0,0) rectangle (#1,#1);
    \end{tikzpicture}
}

\newcommand{\samname}{Monge SAM\xspace}
\newcommand{\shortsamname}{M-SAM\xspace}
\newcommand{\loss}{\ell\left(\boldsymbol{\theta}\right)}
\newcommand{\losspert}{\ell\left(\boldsymbol{\theta} + \boldsymbol{\delta}\right)}
\newcommand{\lgrad}{\nabla \loss}
\newcommand{\lgradnorm}{\lVert\lgrad\rVert_2}

\newcommand{\effpert}[1]{\tilde{\rho}_{\text{#1}}}

\newcommand{\fisher}{\mathbf{F}\left(\boldsymbol{\theta}\right)}
\newcommand{\G}{\mathbf{G}\left(\boldsymbol{\theta}\right)}
\newcommand{\A}{\mathbf{A}_{\rho}\left(\boldsymbol{\theta}\right)}

\icmltitlerunning{\samname: Robust Reparameterization-Invariant Sharpness-Aware Minimization Based on Loss Geometry}

\begin{document}

\twocolumn[
\icmltitle{\samname: Robust Reparameterization-Invariant \\ Sharpness-Aware Minimization Based on Loss Geometry}



\icmlsetsymbol{equal}{*}

\begin{icmlauthorlist}
% \icmlauthor{Firstname1 Lastname1}{equal,yyy}
\icmlauthor{Albert Kjøller Jacobsen}{dtu}
\icmlauthor{Georgios Arvanitidis}{dtu}
% \icmlauthor{Firstname3 Lastname3}{comp}
\end{icmlauthorlist}

\icmlaffiliation{dtu}{Section for Cognitive Systems, DTU Compute, Technical University of Denmark, Kongens Lyngby, Denmark}

\icmlcorrespondingauthor{Albert Kjøller Jacobsen}{akjja@dtu.dk}

\icmlkeywords{Sharpness-Aware Minimization, Deep Learning, Optimization, Differential Geometry, ICML}

\vskip 0.3in
]



\printAffiliationsAndNotice{}  % leave blank if no need to mention equal contribution

\begin{abstract}
Recent studies on deep neural networks show that flat minima of the loss landscape correlate with improved generalization.
Sharpness-aware minimization (SAM) efficiently finds flat regions by updating the parameters according to the gradient at an adversarial perturbation. 
The perturbation depends on the Euclidean metric, making SAM non-invariant under reparametrizations, which blurs sharpness and generalization.
We propose \samname (\shortsamname), a reparametrization invariant version of SAM by considering a Riemannian metric in the parameter space induced naturally by the loss surface.
Compared to previous approaches, \shortsamname works under any modeling choice, relies only on mild assumptions while being as computationally efficient as SAM.
We theoretically argue that \shortsamname varies between SAM and gradient descent (GD), which increases robustness to hyperparameter selection and reduces attraction to suboptimal equilibria like saddle points.
We demonstrate this behavior both theoretically and empirically on a multi-modal representation alignment task. 

% \textbf{Code:} \hfill \href{https://github.com/albertkjoller/sharpness-aware-minimizers}{\texttt{sharpness-aware-minimizers}}

% Recent studies on deep neural networks reveal how flat parameter space regions correlate with improved generalization. Sharpness-aware minimization (SAM) has been proposed for efficiently finding flat regions by updating parameters according to the gradient at an adversarial perturbation, however, SAM's notion of sharpness depends on the network's parameterization, blurring the relation between sharpness and generalization. Other previous works propose reparameterization-invariant versions of SAM, each having their respective drawbacks; one requires probabilistic models while another requires predefining the manifold on which to optimize. We propose an adjusted reparameterization-invariant version pf SAM - called \samname - by introducing a loss-awareness in the perturbation via the Monge metric. This works for any model formulation and loss function while being as computationally efficient as SAM. We theoretically argue that \samname is conservative, resulting in trading off SAM behavior with gradient descent (GD) behavior. This eventually increases robustness to hyperparameter selection and reduces attraction to suboptimal equilibria like saddle points. We demonstrate this theoretically and also empirically on a multi-modal representational alignment task.
\end{abstract}

\section{Introduction}
\label{introduction}

\begin{figure}[t]
    \centering
    \includegraphics[width=\linewidth]{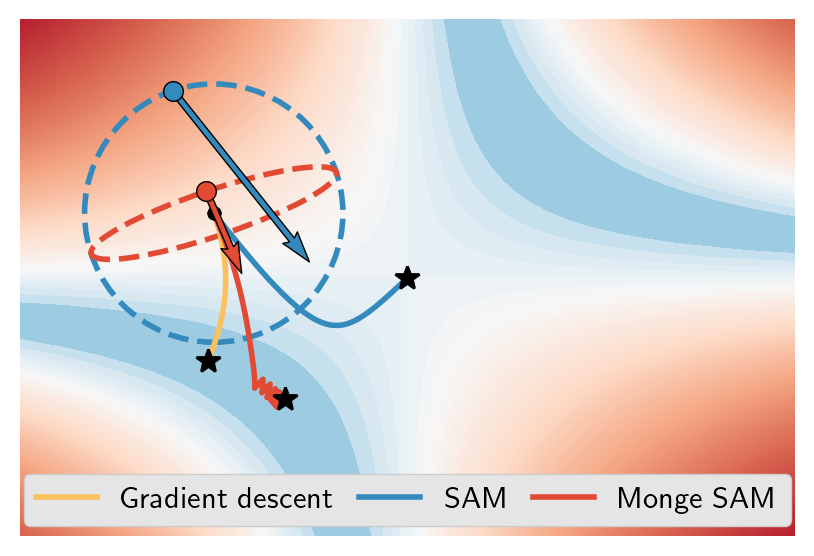}
    \vspace{-15pt}
    \caption{The SAM finds the adversarial perturbation within a Euclidean ball (\dottedline[cyan]{0.125cm} ) which upper bounds the \shortsamname perturbation that is based on the local geometry of the loss (\dottedline[red]{0.125cm} ), implying an adaptive trade-off between SAM and GD. In a loss defined by $\ell\left(\boldsymbol{\theta}\right) = \left(1-\theta_1\theta_2\right)^2$ with banana-shaped minima at $\theta_1\!=\!{1}/{\theta_2}$, \shortsamname is less prone to get attracted to the saddle point at $\boldsymbol{\theta}_s = \left(0,0\right)$ than SAM. \shortsamname can reach lower losses like GD while being capable of walking along minima, eventually finding the flattest global minimum at $\boldsymbol{\theta}^\ast_{\text{flat}} = \left(-1,-1\right)$. We run 200 steps from $\boldsymbol{\theta}_0=(-\frac{3}{2}, \frac{1}{2})$ with $\rho=1$ and a learning rate of $0.01$. Arrows represent the respective gradients (rescaled) at the perturbed points.}
    \label{fig:figure1}
\end{figure}

What makes overparameterized deep neural networks capable of generalizing as well as they do? Inspired by the early work of \citet{hochreiter1997flat} that argued how \emph{flat} regions of the parameter space correspond to networks with low expected overfitting, recent works showed correlation between flatness and generalization capabilities \cite{keskar2016large, dziugaite2017computing, izmailov2018averaging, jastrzkebski2018finding,
chaudhari2019entropy, jiang2019fantastic}, including downstream performance of pre-trained large language models (LLMs) \cite{liu2023same}.

The sharpness-aware minimization (SAM) approach to optimization \cite{foret2020sharpness} has increasingly gained traction due to its practical formulation that seeks flat parameter space regions without requiring information about the curvature through the Hessian, as it only relies on first-order derivatives. Though SAM might enhance generalization, it is known that sharp minima can also generalize equivalently whenever the learned function is reparameterized accordingly \cite{dinh2017sharp, kristiadi2024geometry}. Therefore, several extensions to SAM have been proposed; \citet{kwon2021asam} proposed adaptive SAM (ASAM) to partly tackle this reparameterization issue by making SAM scale-invariant, where Fisher SAM \cite{kim2022fisher} adapts the perturbation step of SAM by employing a Riemannian metric induced in the parameter space under the Information Geometry framework, thereby respecting the local geometry of probabilistic models. Similarly, Riemannian SAM \cite{yun2024riemannian} generalizes this framework, which includes Fisher SAM, by considering the parameter space of a deep neural network as a predefined Riemannian manifold, as a sphere, and computing perturbations and gradient updates on the manifold.

% In contrast, we propose a more general approach to making SAM reparameterization-invariant that does not rely on a probabilistic formulation or predefined manifolds while still respecting the geometry of the loss surface.

% Our contributions
In this paper, we propose a novel approach to tackle the reparameterization issue of SAM. 
We naturally induce a Riemannian metric in the parameter space that captures the geometry of the loss surface, namely the Monge metric, and thereby constrain the region in which the adversarial SAM perturbation is searched for.
Our approach is more general than previously proposed approaches as it does not rely on a probabilistic formulation or predefined manifolds while still respecting the local structure of the loss. Due to the simplicity of this metric, our proposed method has an analytical expression of the adversarial perturbation, in contrast to Fisher SAM,  which typically requires regularizing and approximating the metric through diagonalization.

Our main contributions include the following:
\begin{enumerate}
    \item We establish \emph{\samname (\shortsamname)}; a novel sharpness-aware minimizer that exploits the Monge metric.
    % a closed-form solution for the approximate worst-case perturbation.
    \item \shortsamname relies on mild assumptions and is not based on probabilistic model formulations or predefining the parameter manifold; it works for any modeling choice.    
    \item We theoretically justify that \shortsamname is less prone to get attracted to suboptimal equilibria like saddle points. We extend on previous studies by analyzing stability of the SAM gradient flow using perturbation theory.
    \item We provide empirical evidence of \shortsamname being more robust to hyperparameter selection than SAM, having a beneficial impact on performance for some tasks.
    % \item Fine-tuning with \shortsamname increases representational alignment of the multi-modal CLIP model by forming more distinct conceptual regions.
\end{enumerate}

\section{Background}

\paragraph{Notation.} We denote the $K$-dimensional parameter space of a parametric data-dependent model, $f_{\boldsymbol{\theta}}:\mathbb{R}^D \rightarrow\mathbb{R}^C$ (e.g. a neural network) by $\boldsymbol{\theta} = \left(\theta_1, \dots, \theta_K\right) \in \Theta \subseteq \mathbb{R}^K$. We define general loss functions as $\ell: \Theta \times \mathcal{X} \times \mathcal{Y} \rightarrow \mathbb{R} 
$ where $\mathcal{X}\in\mathbb{R}^D$ and $\mathcal{Y}\in\mathbb{R}^C$ are the input and output spaces, respectively. For simplicity we disregard the dependency on the data $\left(\boldsymbol{x}, \boldsymbol{y}\right) \in \mathcal{X} \times \mathcal{Y}$ and denote the general loss function by $\ell\left(\boldsymbol{\theta}\right)$ for a specific realization of the model $f_{\boldsymbol{\theta}}$.

\subsection{Sharpness-Aware Minimization}

As mentioned, SAM \cite{foret2020sharpness} seeks flat minima without relying on Hessian-based measures as such would be impractical to compute. SAM seeks to minimize the loss at a perturbed parameter set, $\boldsymbol{\theta} + \boldsymbol{\delta}$, where $\boldsymbol{\delta}$ is the adversarial perturbation. Thus SAM's optimization objective is:
\begin{equation}
    \underset{\boldsymbol{\theta}}{\min} \     \underset{\lVert\boldsymbol{\delta} \rVert_{\mathbf{M}} \leq \rho}{\max} \ \losspert,
\end{equation}
where $\lVert\boldsymbol{\delta} \rVert^2_{\mathbf{M}} = \boldsymbol{\delta}^\top \mathbf{M}\boldsymbol{\delta}$ defines the local norm under the positive-definite matrix $\mathbf{M}$.

The type of flatness minimized by SAM appears in the objective by adding and subtracting the regular loss term and rearranging gives:
\begin{equation}
    \underset{\boldsymbol{\theta}}{\min} \overset{\text{Loss term}}{\overbrace{\loss}} + \ \overset{\text{Sharpness term}}{\overbrace{\underset{\lVert\boldsymbol{\delta} \rVert_{\mathbf{M}} \leq \rho}{\max} \ \losspert - \loss}} 
\end{equation}
Thus, SAM's notion of sharpness is the difference in loss values evaluated at the original parameters and the worst-case perturbation of the parameters, under the local norm constraint.
%taken within a $p$-ball with radius $\rho$. 
In practice, finding the worst-case perturbation is handled by solving the dual-norm problem that occurs when approximating the objective with a first-order Taylor expansion. Previous works consider $\mathbf{M}=\mathbb{I}_K$ as the default choice, i.e. searching withing the Euclidean ball, for which the worst-case perturbation is the rescaled gradient:
\begin{equation}
    \boldsymbol{\delta}^\ast_{\text{SAM}} = \frac{\rho}{\lgradnorm} \cdot\lgrad.
\end{equation}
After computing the worst-case perturbation, minimizing the SAM objective requires an additional backward pass under the \emph{base optimizer} (e.g. stochastic gradient descent). 
Remark that the scaling factor $\effpert{SAM} := \rho \ / \ \lgradnorm$ will later be referred to as the \emph{effective perturbation size}. 

% After computing the worst-case perturbation, minimizing the SAM objective is straight-forward and only requires defining the \textit{base optimizer} to use (e.g. stochastic gradient descent), thus introducing an additional backward pass. 
% Also, we add a tiny value to the denominator in practice. Remark that the scaling factor will later be referred to as the \textit{effective perturbation size}, denoted by $\effpert{SAM} := \rho \ / \ \lgradnorm$. 

\subsection{Fisher SAM}

Fisher SAM \cite{kim2022fisher} exploits the fact that the KL-divergence between two infinitesimally close parametric distributions can be approximated by the norm under a Riemannian metric. In particular, an approximation to this metric is the empirical Fisher information matrix. 
% Fisher SAM \cite{kim2022fisher} exploits the fact that the KL-divergence is locally a distance function between distributions and a Riemannian metric that approximately equals the empirical Fisher information matrix, also called the Fisher metric, $\fisher$. 
Leveraging this fact, the optimization objective is rephrased as 
\begin{equation}
    \underset{\boldsymbol{\theta}}{\min} \     \underset{\boldsymbol{\delta}^\top \fisher \boldsymbol{\delta} \leq \rho^2}{\max} \ \losspert.
\end{equation}
Approximating this with a first-order Taylor expansion leads to a quadratically constrained linear program for the worst-case perturbation. Solving the Lagrangian gives:
\begin{equation}
    \label{eq:fishersam}\boldsymbol{\delta}_{\text{Fisher}}^\ast = \rho\cdot\frac{\fisher^{-1} \lgrad}{\sqrt{\lgrad \fisher^{-1}\lgrad}}.
\end{equation}
Fisher SAM is reparameterization-invariant and respects the local geometry of the probabilistic model through the Fisher metric. The Fisher metric is defined as the sum of the outer products of the gradient of log-likelihoods, $\mathbf{F}\left(\boldsymbol{\theta}\right) = \mathbb{E}_{\boldsymbol{x}}\mathbb{E}_{\boldsymbol{\theta}} [\nabla \log p\left(y|\boldsymbol{x}, \boldsymbol{\theta}\right) \nabla \log p\left(y|\boldsymbol{x}, \boldsymbol{\theta}\right)^\top]$, which grows quadratically with the number of parameters, and reveals that Fisher SAM only works for probabilistic models. In practice, the implementation relies on approximation by mini-batching and diagonalization as $\hat{\mathbf{F}}\left(\boldsymbol{\theta}\right)=\operatorname{diag}\left[\frac{1}{|\mathcal{S}|}\sum_{i\in\mathcal{S}} \nabla \log p\left(y_i|\boldsymbol{x}_i, \boldsymbol{\theta}\right)\right]^2$ with $\mathcal{S}$ being a batch. This ad-hoc approximation is necessary as inverting the original metric is prohibitively expensive, while to avoid division by zero the inverse is defined as $\hat{f}_i^{-1} = 1/(1+\eta f_i)$ where $f_i$ denotes the diagonal elements of $\hat{\mathbf{F}}\left(\boldsymbol{\theta}\right)$ and $\eta$ a hyperparameter. Note that diagonalization removes correlation, which underrepresents the local structure.
% limiting the practical applications of Fisher SAM, e.g. for LLM training.

\section{\samname}
We develop a geometry-aware version of SAM that is not based on probabilistic model formulations or predefined parameter manifolds. Intuitively, we approach this problem by modifying the parameter space $\Theta$ to locally be aware of the training loss structure. 
We propose equipping the parameter space with a simple Riemannian metric that encodes the loss surface geometry, which is computationally efficient and enables us to obtain a closed-form expression of the worst-case perturbation. 

% We propose equipping the parameter space with a simpler Riemannian metric than the Fisher metric which is computationally efficient, does not require a probabilistic formulation, and enables us to obtain a closed-form expression of the worst-case perturbation. 
% We name the adjusted approach as \samname, abbreviated \shortsamname.

\subsection{The Monge metric} 
% \paragraph{Monge metric.}
Given the definition of a general loss function $\ell(\boldsymbol{\theta})$, we consider the loss surface of the model $f_{\boldsymbol{\theta}}$ as a $K$-dimensional smooth manifold embedded in $\mathbb{R}^{K+1}$ as follows $\mathcal{M} = g\left(\boldsymbol{\theta}\right)= \left[\theta_1, \dots, \theta_K, \loss\right] \in \mathbb{R}^{d+1}$. 
The parameter space $\Theta$ represents the \emph{intrinsic coordinates} of the manifold $\mathcal{M}$, and at a point $g\left(\boldsymbol{\theta}\right)\in \mathcal{M}$ the tangent space $\mathcal{T}_{g\left(\boldsymbol{\theta}\right)}\mathcal{M}$ is spanned by the Jacobian $\mathbf{J}_g:\Theta \rightarrow\mathbb{R}^{K+1 \times K}$.
We can therefore write a tangent vector to the manifold in terms of its intrinsic coordinates $\vec{\boldsymbol{v}} \in \mathbb{R}^K$ as $\mathbf{J}_g\left(\boldsymbol{\theta}\right) \vec{\boldsymbol{v}}$. 
Likewise, the inner product between two tangential vectors, $\vec{\boldsymbol{v}}_1$ and $\vec{\boldsymbol{v}}_2$, in the same tangent plane is $\langle \mathbf{J}_g\left(\boldsymbol{\theta}\right) \vec{\boldsymbol{v}}_1, \mathbf{J}_g\left(\boldsymbol{\theta}\right) \vec{\boldsymbol{v}}_2 \rangle = \vec{\boldsymbol{v}}_1^\top \mathbf{J}_g\left(\boldsymbol{\theta}\right)^\top \mathbf{J}_g\left(\boldsymbol{\theta}\right) \vec{\boldsymbol{v}}_2$. Evidently, the induced \emph{Riemannian metric} $\G = \mathbf{J}_g\left(\boldsymbol{\theta}\right)^\top\mathbf{J}_g\left(\boldsymbol{\theta}\right)$ provides a notion of local inner products on the manifold through its intrinsic coordinates and captures the local geometry.

Due to the parametrization $g(\boldsymbol{\theta})$ of the manifold $\mathcal{M}$ the Jacobian takes a simple form, i.e. $\mathbf{J}_g\left(\boldsymbol{\theta}\right) = \left[\mathbb{I}_K,\lgrad \right]^\top$. Consequently, this gives a simple expression for the metric:
\begin{equation}
    \G = \mathbb{I}_K + \lgrad \lgrad^\top,
\end{equation}
which is known as the \emph{Monge} metric and consists of the outer product of the loss gradient with itself, regularized by adding 1's to the diagonal entries. The inherent regularization ensures positive definiteness of $\G$ and thereby invertibility. Following the Sherman-Morrison formula the inverse metric takes the form:
\begin{equation}
    % \G^{-1} = \mathbf{I}_K - \frac{\lgrad \lgrad^\top}{1 + \lgrad^\top \lgrad}
    \G^{-1} = \mathbb{I}_K - \frac{\lgrad \lgrad^\top}{1 + \lgradnorm^2}.
\end{equation}

Using the Monge metric rather than the Fisher metric has several advantages; while gradients of the log-likelihood can indeed be obtained for probabilistic models, accessing these quantities requires careful implementation when relying on automatic differentiation frameworks. Contrarily, the gradient of the loss is straight-forward to obtain for any implementation. Secondly, the inherent regularization ensures a closed-form expression of the inverse Monge metric, thereby allowing us to use its full expressivity without using tricks like diagonalization and explicit regularization. 

\subsection{The \samname perturbation}

Leveraging the notion of local inner products as expressed using the Monge metric, we slightly change the objective:
\begin{equation}    \underset{\boldsymbol{\theta}}{\min} \     \underset{\boldsymbol{\delta}^\top \G \boldsymbol{\delta} \leq \rho^2}{\max} \ \losspert.
\end{equation}
Though solving for the worst-case perturbation gives a similar expression as in Equation \ref{eq:fishersam}, this further simplifies due to the inverse Monge metric being analytically accessible, resulting in a closed-form expression for \shortsamname's worst-case perturbation:
\begin{equation}
    \boldsymbol{\delta}_{\text{\shortsamname}}^\ast = \frac{1}{\sqrt{1 + \lgradnorm^2}} \ \cdot \underset{= \ \boldsymbol{\delta}_{\text{SAM}}^\ast}{\underbrace{ \ \frac{\rho}{\lgradnorm} \cdot \lgrad}}.
\end{equation}
We see that the geometry-aware worst-case perturbation is a rescaled version of SAM's worst-case perturbation and define $\effpert{\shortsamname} := \rho \ / \ ( \ \lgradnorm \cdot \sqrt{1 + \lgradnorm^2} \ )$  as \shortsamname's effective perturbation radius. As such, the computational requirements are comparable to SAM.

\begin{figure*}[t!]
    \raggedright
    \includegraphics[width=\linewidth]{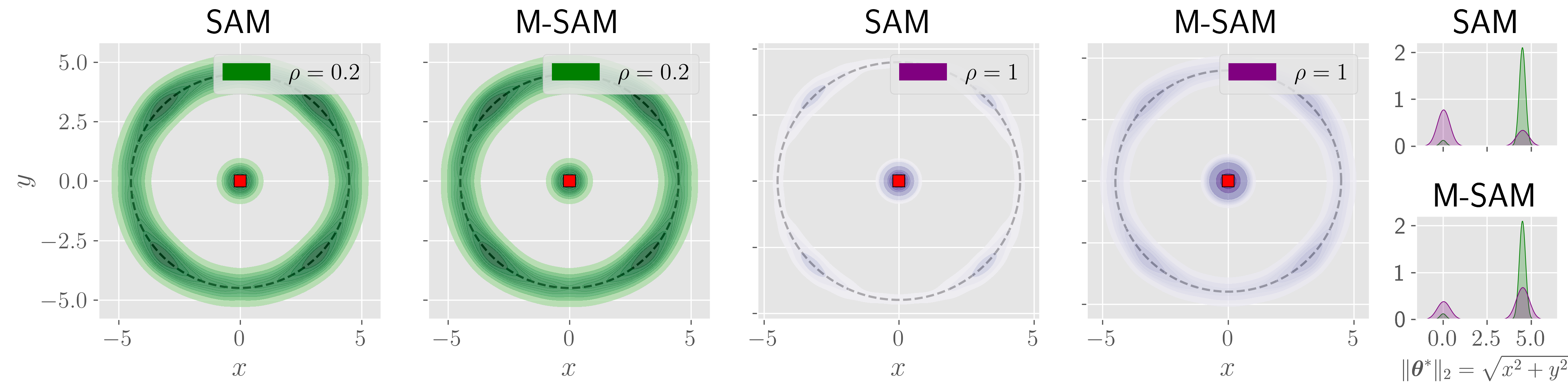}
    % \includegraphics[width=0.45\linewidth]{figures/2d-sinc-xy-dist-sam.png}
    % \hfill
    % \includegraphics[width=0.45\linewidth]{figures/2d-sinc-xy-dist-msam.png}
    % \hspace{0.4cm}
    \vspace{-18pt}
    \caption{\textbf{Attraction to maxima.} We consider the scaled 2D $\mathrm{sinc}$-function given by $\mathrm{sinc}\left(x, y\right) = 5\cdot\sin \left(x^2+y^2\right) / \left(x^2+y^2\right)$ and draw $N=200$ samples of $\boldsymbol{\theta}=(x,y)$, uniformly distributed within the centered unit square. For each sample, we run SAM and \shortsamname for 200 steps with a learning rate of 0.05 and $\rho \in \{0.2, 1\}$ and plot the distribution of the converged estimates, $\boldsymbol{\theta}^\ast$. The larger perturbation radius $\rho=1$ makes the global maximum ( \redsquare{0.18}) at $\boldsymbol{\theta}=(0,0)$ a stronger attractor for SAM, as for $\rho=0.2$ SAM is more likely to descend into the surrounding circular minima range (\dottedline[black]{0.125cm} ). We observe similar trends for \shortsamname, yet see that the conservative property restricts how strong attractor the maxima is, even for the high perturbation radius of $\rho=1$.}
    \label{fig:2d-sinc}
\end{figure*}
\begin{figure}[t!]
    \raggedright
    \includegraphics[width=0.96\linewidth]{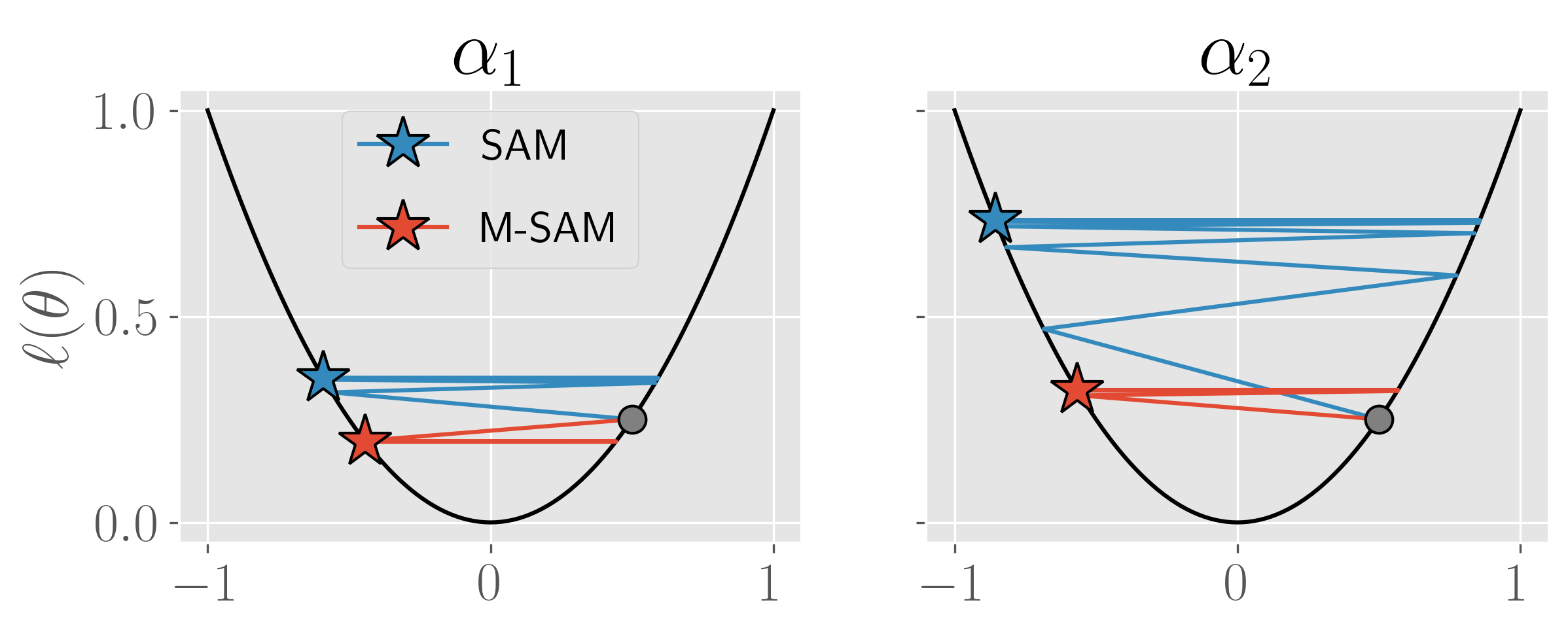}
    \vspace{-10pt}
    \caption{\textbf{Conservative property.} SAM vs. \shortsamname behavior in a simple paraboloid loss given by $\ell\left(\theta\right) = \theta^2$ for two learning rates, $\alpha_1 < \alpha_2$ when fixing the perturbation radius $\rho=0.3$ and taking $11$ steps. \shortsamname always converges to lower values than SAM and is additionally limitedly affected by large learning rates, revealing \shortsamname's conservativeness that originates from being loss-aware.}
    \label{fig:learning-rates}
\end{figure}

\paragraph{\samname is conservative.}
We now consider how the adjusted worst-case perturbation behaves under two extreme cases, namely at locations in the parameter space where the loss is either 1) very steep or 2) close to being a stationary point, i.e. $\lgradnorm \rightarrow+ \infty$ and $\lgradnorm \rightarrow 0$, respectively. The asymptotic behavior is:
\begin{alignat*}{3}
    &\underset{\lgradnorm \ \rightarrow \ + \infty}{\lim} \ &&\boldsymbol{\delta}_{\text{\shortsamname}}^\ast &{}={}& \boldsymbol{0} \\
    &\underset{\lgradnorm \ \rightarrow \ 0}{\lim} \ &&\boldsymbol{\delta}_{\text{\shortsamname}}^\ast &{}={}& \boldsymbol{\delta}_{\text{SAM}}^\ast.
\end{alignat*}
Put differently, the effective perturbation radius of \shortsamname is upper bounded by that of SAM, i.e. $\tilde{\rho}_{\text{\shortsamname}} \leq \tilde{\rho}_{\text{SAM}}$. Remarkably, \shortsamname appears to \emph{trade-off} between gradient descent (GD) and SAM behavior by restricting the perturbation inversely to the growth of the training loss where the original SAM perturbation can in principle explode. 

In Figure \ref{fig:learning-rates} we show how \shortsamname is biased towards GD behavior. We apply SAM and \shortsamname in a symmetric 1D toy loss defined by a second-order polynomial, i.e. $\ell\left(\theta\right)=\theta^2$. We compare trajectories of the two methods for two learning rates $\alpha_1=0.66$ and $\alpha_2=0.74$ when fixing the perturbation radius to $\rho=0.3$. For $\alpha_1$ the two methods behave differently; the SAM perturbation - and consequently also the gradient update - is sufficiently large for SAM to diverge from the initial location whereas \shortsamname has a lower effective perturbation radius and thus take smaller gradient updates, thereby initially taking converging steps and converging lower in the loss. Though both methods diverge for $\alpha_2$, \shortsamname converges remarkably lower in the loss. As \shortsamname favors exploiting the current valley rather than exploring the parameter space through larger perturbations, we argue that \shortsamname is \emph{conservative} compared to SAM.

\paragraph{Reduced attraction to saddle points.}
\label{section:saddle}
We now study a property of \shortsamname that originates from its conservativeness. SAM has the undesirable property of getting attracted to saddle points \cite{kaddour2022flat, compagnoni2023sde, kim2023stability}. Less discussed is the fact that this attraction holds for any type of suboptimal equilibria, meaning that SAM \emph{can} get attracted to maxima as well \cite{ujvary2022rethinking, compagnoni2023sde}. We present an example in Figure \ref{fig:2d-sinc}, where we also show that \shortsamname is less prone to get attracted to such equilibria.

\begin{figure*}[t]
\centering
    \includegraphics[width=0.8\linewidth]{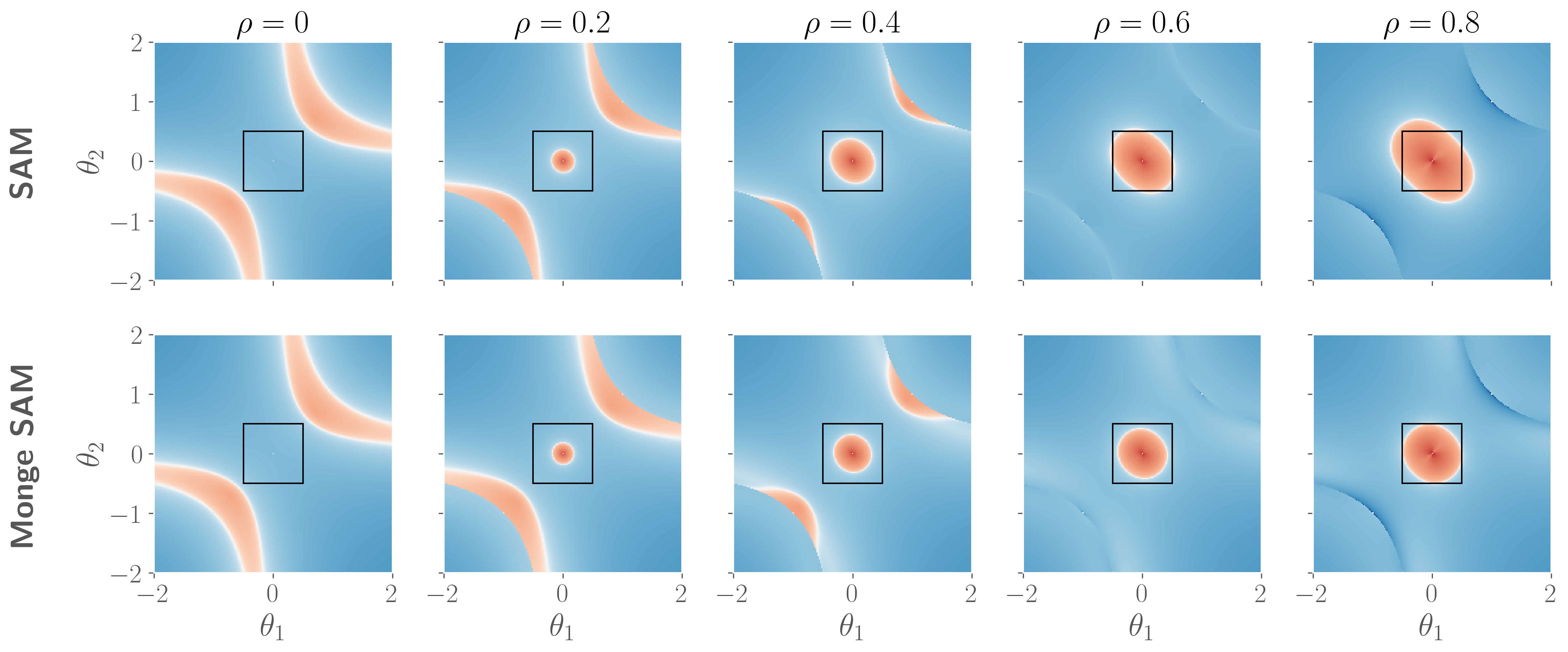}
    \hspace{1.1cm}
    \vspace{-10pt}
    % \raggedright
    % \includegraphics[width=0.92\linewidth]{figures/saddle-points.png}
    \caption{\textbf{Stability criteria.} Intuitively, we highlight the stability of a parameter $\boldsymbol{\theta}$ by considering the behavior of a random sample $\tilde{\boldsymbol{\theta}} = \boldsymbol{\theta} + \boldsymbol{\epsilon}$ within the $\boldsymbol{\epsilon}$-neighborhood of $\boldsymbol{\theta}$.
    %, $\tilde{\boldsymbol{\theta}} = \boldsymbol{\theta} + \boldsymbol{\epsilon}$, within the $\boldsymbol{\epsilon}$-neighborhood of $\boldsymbol{\theta}$. 
    We consider the banana-shaped loss, i.e. $\loss = \left(1-\theta_1\theta_2\right)^2$, and compute the eigenvalues, $\{\lambda_1, \lambda_2\}$, of $\A$. We show $\lambda_2$ across the parameter space for varying perturbation radii for SAM and \shortsamname. As $\lambda_1$ is always positive, stability of SAM and \shortsamname dynamics is satisfied when $\lambda_2$ is positive (red regions). The square is used for reference. When $\rho=0$, i.e. GD behavior, the $\boldsymbol{\epsilon}$-neighborhood of any point in parameter space will converge to the global minima, revealing that GD does not suffer from saddle point attraction if not initialized exactly at the saddle point, $\boldsymbol{\theta}_s=\left(0,0\right)$.
    % the global minima are the only stable regions, revealing that GD does not suffer from saddle point attraction. 
    Contrarily, $\boldsymbol{\theta}_s$ is stable for SAM and \shortsamname for higher $\rho$, 
    even so, \shortsamname's region of attraction is smaller near the saddle but larger near flat global minima at $\boldsymbol{\theta}_{\text{flat}}^\ast \in \{(-1, -1), (1,1)\}$, due to its conservative nature.}
    \label{fig:saddle-points}
\end{figure*}

We consider the close neighborhood of any parameter configuration $\boldsymbol{\theta}$ and define it by a Gaussian with sufficiently low variance $\sigma^2$, i.e. $\tilde{\boldsymbol{\theta}} \sim \mathcal{N}\left(\boldsymbol{\theta}, \sigma^2\mathbb{I}_K\right)$. For simplifying expressions, we reparameterize the neighborhood as $\tilde{\boldsymbol{\theta}} = \boldsymbol{\theta} + \boldsymbol{\epsilon}$ where $\boldsymbol{\epsilon} \sim \mathcal{N}\left(\boldsymbol{0}, \sigma^2\mathbb{I}_K\right)$. With the aim of examining attraction to suboptimal equilibria, we describe the general SAM dynamics at $\tilde{\boldsymbol{\theta}}$ using SAM's gradient flow, eventually giving rise to an ordinary differential equation (ODE) governing the neighborhood:
\begin{equation}
    \frac{d\boldsymbol{\epsilon}}{dt} \approx - \ \underset{\mathbf{A}_{\rho}\left(\boldsymbol{\theta}\right)}{\underbrace{\biggl(\nabla^2 \ell\left(\boldsymbol{\theta} + \boldsymbol{\delta}^\ast\right) \bigl[ \mathbb{I}_K + \tilde{\rho} \cdot \nabla^2 \ell \left(\boldsymbol{\theta}\right) \bigr]\biggr)}}^\top \boldsymbol{\epsilon}.
\end{equation}
We provide the associated proof in 
Appendix \ref{appendix:proof-ODE}.
Here, $\tilde{\rho}$ is the general effective perturbation radius and could be replaced by $\tilde{\rho}_{\text{SAM}}$ or $\tilde{\rho}_{\text{\shortsamname}}$. In essence, the ODE describes how a random sample within the close neighborhood of $\boldsymbol{\theta}$ behaves, yet further analysis of the dynamics described by $\A$ is essential for understanding SAM's attraction to suboptimal equilibria. First, we remark the minus-sign, revealing that directions in which $\A$ expands, i.e. the directions with positive eigenvalues, instead pull nearby points towards $\boldsymbol{\theta}$. Hence, the point $\boldsymbol{\theta}$ is an \emph{attractor} if all eigenvalues of $\A$ are positive. Though the eigenvalues of $\A$ do not have an analytical expression for arbitrary $\boldsymbol{\theta}$, the dynamics simplify remarkably close to equilibria since $\lgradnorm \approx 0$, hence $\boldsymbol{\delta}^\ast \approx \boldsymbol{0}$. As argued by \citet{kim2023stability} this gives rise to a criteria of stability at equilibria under SAM dynamics depending on the local curvature:
\begin{equation}
    \tilde{\rho} > -{1}/{\lambda_i}, \quad \forall \lambda_i \in \{\lambda_1, \dots,\lambda_K\},
\end{equation}
where $\{\lambda_1, \dots, \lambda_K\}$ are the eigenvalues of the Hessian, $\nabla^2\loss$. The criteria reveals that \emph{any equilibria can be stable} under SAM dynamics if the perturbation is relatively large compared to the local curvature, e.g. if $\tilde{\rho} = 0.35$ and we consider a saddle point or a maximum with eigenvalues of $\{-3, 3\}$ and $\{-3, -3\}$, respectively. This aligns with Figure \ref{fig:2d-sinc} where the maximum of the $\operatorname{sinc}$-function is a stronger attractor when the perturbation radius is large. 

The direct interpretation of \shortsamname's conservativeness, namely that $\tilde{\rho}_{\text{\shortsamname}} \leq \tilde{\rho}_{\text{SAM}}$, reveals that the proposed stability criteria is less frequently satisfied for \shortsamname since $\tilde{\rho}_{\text{SAM}} \geq \tilde{\rho}_{\text{\shortsamname}} > -{1}/{\lambda_i}$, and hence \emph{\shortsamname is attracted less to suboptimal equilibria than SAM}. We demonstrate empirically this beneficial property of \shortsamname in Figure \ref{fig:saddle-points} by computing the eigenvalues of $\A$ across the space for the function that we considered in Figure \ref{fig:figure1}. Eigenvalue positivity serves as a criteria for stability and we see that SAM's region of attraction is larger than \shortsamname's around the saddle point for larger perturbation radii. The behavior is the opposite when close to flat global minima.

\begin{figure*}
    \centering
    \includegraphics[width=0.95\linewidth]{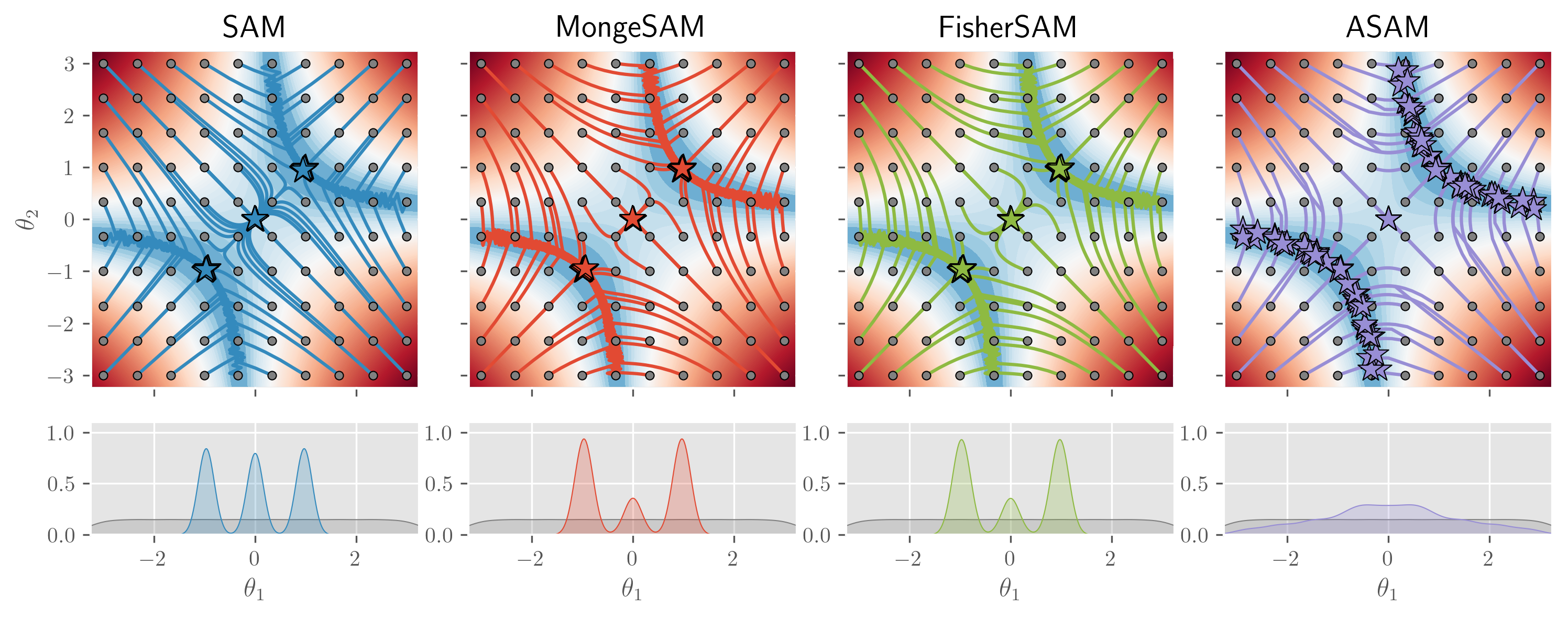}
    \vspace{-10pt}    \caption{\textbf{Comparing sharpness-aware minimizers.} Trajectories of SAM, \shortsamname, Fisher SAM and ASAM in the banana-shaped loss given by $\loss = (1-\theta_1\theta_2)^2$. Each optimizer is initialized on the points of a $10\times10$ grid and runs to convergence. The marginal distributions of convergence locations summarize the final estimates. While all methods can explore the valley of minima and locate the flattest solution, SAM exhibits a notable tendency to converge to the saddle point. \shortsamname and Fisher SAM show qualitatively similar behavior, with only minor differences when initialized near the saddle. We note that this behavior is due to the diagonal approximation of the Fisher metric. We compare with ASAM for completeness; it mostly avoids the saddle point but converges anywhere on the minima range and does generally not find one of the flattest solution.}
    \label{fig:comparison}
\end{figure*}

% \subsection{Toy Example in 2D: revisiting Figure \ref{fig:figure1}}
\subsection{Illustrating \samname: Toy Example in 2D}
In Figure \ref{fig:comparison} we consider the loss $\loss =\left(1-\theta_1\theta_2\right)^2$ with banana-shaped minima (as in Figure \ref{fig:figure1}). We initialize SAM, ASAM and \shortsamname on a uniform grid $(\theta_1, \theta_2) \in [-3, 3] \times [-3, 3]$ and run the optimizers from each point for 200 steps using a learning rate of $0.01$ and a large perturbation radius of $\rho=1$. We remark the non-probabilistic formulation, yet we compare the two approaches to Fisher SAM. 
% We also compare with ASAM for completeness.

All methods are clearly capable of finding the flatter solution when reaching the valleys of global minima; a key property of SAM that motivates its use for fine-tuning tasks and originates from normalizing the gradient in the perturbation step \cite{dai2024crucial}. Yet, \shortsamname reaches the flat global minima more easily than SAM which gets attracted to the saddle point, even when initialized far from it. Fisher SAM appears to be equally robust as \shortsamname in this manner and they also behave remarkably similar when considering their marginal distributions of convergence. Only a few of the trajectories close to the center qualitatively differ, which reflects the implications of the ad-hoc approximation of the inverse Fisher metric with a diagonal matrix where \shortsamname is equally fast but also captures the correlation.

\section{Experiments}
We empirically demonstrate how \shortsamname behaves in the context of deep neural networks. We rely on mini-batches which introduces stochasticity in the method, whereas the analysis above is based on the deterministic version. We compare with SGD and SAM for understanding \shortsamname in high-dimensional settings and include a comparison with Adam on a representational alignment task.

\subsection{Fine-tuning from a Bad Global Minimum}

\begin{figure}[t!]
    \raggedright
    \includegraphics[width=0.98\linewidth]{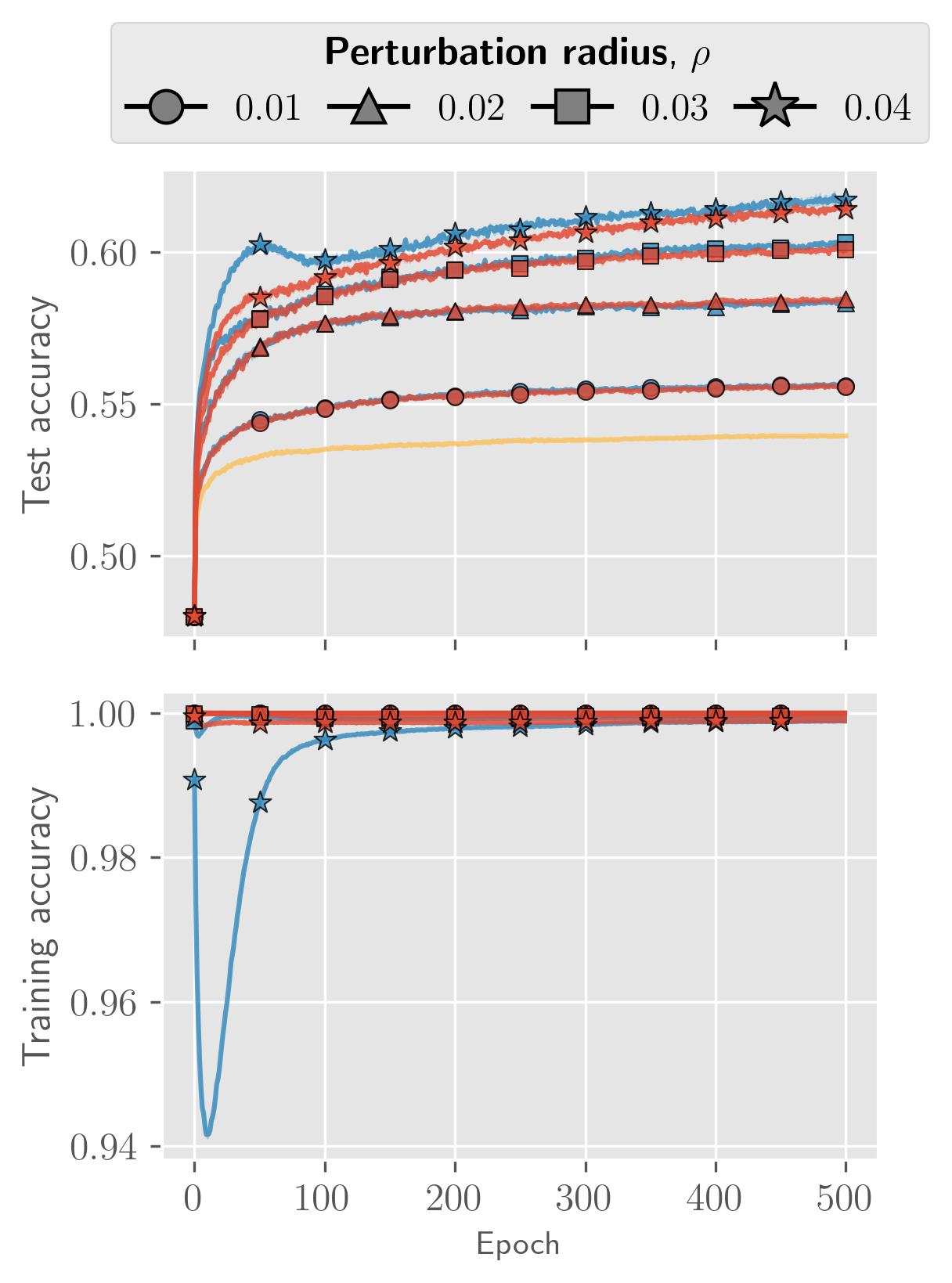}
    \vspace{-10pt}
    \caption{\textbf{Fine-tuning a bad ResNet-18.} We initialize an adversarially trained ResNet-18 \cite{liu2020bad} using SGD (\solidline[orange]{0.3cm}), SAM (\solidline[cyan]{0.3cm}) and \shortsamname (\solidline[red]{0.3cm}) with a learning rate set to 0.001. We examine SAM and \shortsamname behavior for varying perturbation radii and see that \shortsamname's conservativeness biases the model to exploit the current valley of global minima rather than escaping and exploring other regions of the parameter space. We provide uncertainty estimates as the standard error of the mean computed from $N=5$ pseudo-random repetitions. Remark that most models perfectly fit the training data, retaining a training accuracy of 100\%.}
    \label{fig:overtraining}
\end{figure}

We consider a pre-trained ResNet-18 for classifying images of CIFAR10, converged to a bad global minima \cite{liu2020bad} that perfectly fits the training data while obtaining 48\% generalization accuracy. We fine-tune with SGD, SAM and \shortsamname using a small learning rate, i.e. $\alpha=0.001$, with the aim of restricting the optimizers to take steps inside the valley containing the bad global minimum, similarly to \citet{dai2024crucial}. We use a fixed batch size of $B=128$ and train for $500$ epochs using a NLL loss on log-softmax outputs of the network. We experiment with $\rho \in \{0.005, 0.01, 0.03, 0.05\}$. Even if some configurations do not converge, we deem the training horizon sufficiently long for examining the dynamics of the methods. The results are presented in Figure \ref{fig:overtraining} with uncertainty estimates empirically computed as the standard error of the mean (SEM) from $N=5$ pseudo-random repetitions.

For lower perturbation radii $\rho \in \{0.005, 0.01, 0.03\}$, SAM and \shortsamname behave similarly as in the 2D toy setting; they are capable of moving along the valley of bad global minima, eventually converging to a solution with increased generalization performance. Specifically, searching wider with $\rho=0.03$ allows for improving approx. $6\%$ compared to fine-tuning with SGD. Though increasing the perturbation size to $\rho=0.05$ increases the model's generalization capabilities even further it also reveals the main controversial property of \shortsamname, namely the  difficulty to escape the valley of bad global minima. Contrarily, SAM with large perturbations (i.e. $\rho=0.05$) quickly escapes the initial valley and finds a region of the parameter space where the generalization performance significantly improves.

We emphasize that this behavior which leads to better performance, in a different scenario might come at the cost of forgetting what was learned during pre-training, potentially making \shortsamname a better choice for fine-tuning. 
As we showed above, large perturbation radii increase SAM's attraction to suboptimal equilibria (Section \ref{section:saddle}); so applying SAM with uncalibrated perturbation parameter $\rho$ to fine-tune pre-trained models that generalize well, potentially in sharp parameter regions, may ultimately result in convergence at another suboptimal location e.g., saddle point. In fact, we provide such an experiment below.

\subsection{How does \shortsamname affect cross-modal alignment?}
A recent study \cite{huh2024platonic} suggested that the alignment of the latent representation space of multi-modal models improves with the capacity of the domain-specific encoders. We examine whether fine-tuning with a sharpness-aware minimizer can improve representational alignment by finding better minima, potentially leading to more robust representations and improved alignment, without requiring increasing the model capacity.

\begin{figure}[t!]
    \raggedright
    \includegraphics[width=\linewidth]{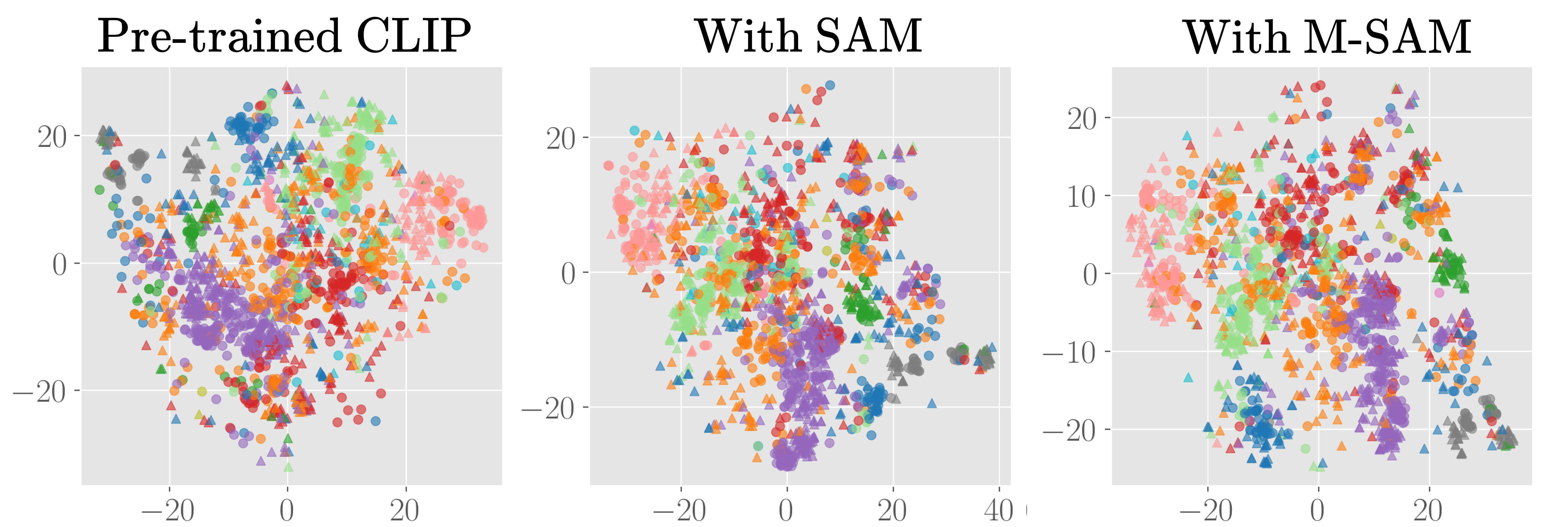}
    \vspace{-15pt}
    \caption{\textbf{Visualizing CLIP space.} We stack the latent text and image representations (MS-COCO) and project them with PCA to a lower-dimensional subspace. The main variation in the stacked representation space originates from the modality. So we remove the information of the first principal component direction before running tSNE on the modified reconstructed latent representations. We label each image-text pair by classifying the images using a ResNet-50, pre-trained on ImageNet labels and color them according to their superclass label (e.g. \textit{Vehicle}, \textit{Clothing} or \textit{Food}), obtained by backtracking the ImageNet graph. We restrict ourselves to consider image-text pairs from the 12 major superclasses.}
    \label{fig:latent-space}
\end{figure}

We fine-tune a transformer-based CLIP model \cite{radford2021learning} on the Wikimedia version of the WIT dataset (available at \textit{huggingface.co}) with SGD, Adam, SAM and \shortsamname using CLIP loss. We used a batch size of 128 and train to convergence. We run a grid-search for SAM and \shortsamname with combinations of learning rates, $\alpha$, and perturbation radii, $\rho$, from $\{1\cdot 10^{-5}, 5\cdot 10^{-5}, 1\cdot 10^{-4}\}$ and $\{0.01, 0.02, 0.03, 0.04\}$, respectively. For SAM we also consider $\rho=0.005$. For SGD and SAM we examine learning rates in the range $[10^{-6}, 10^{-2}]$, yet most settings converge slowly to worse solutions than SAM-based methods or immediately diverge. We evaluate the generalization capabilities of the pre-trained and fine-tuned models on a subset of the MS-COCO Captions dataset \cite{lin2014microsoft}, restricted to having one caption per image. We measure alignment with the mutual $k$NN similarity score, $\mathcal{S}_{k\text{NN}}$, \cite{huh2024platonic} by considering the fraction of  $k=8$ neighbors shared for the latent text and image representation for an image-text input pair; this captures whether conceptual regions form in CLIP space. We report the optimal alignment in Table \ref{tab:alignment} and a comparison of CLIP losses at convergence for SAM and \shortsamname in Table \ref{tab:alignment-fixed-perturbation} when fixing the perturbation radius to $\rho=0.01$. The remaining results are provided in Appendix \ref{appendix:alignment}. We also include a visualization of the latent representations space in Figure \ref{fig:latent-space} for qualitatively assessing the impact of fine-tuning with SAM and \shortsamname.

\begin{table}[t!]
    \centering
    \caption{Alignment scores of fine-tuned CLIP models on MS-COCO. Optimal hyperparameters are shown in Appendix \ref{appendix:alignment}.}
    \vspace{0.25cm}
    \begin{tabular}{l|ccccc}
        \toprule 
        & \textbf{None} & \textbf{SGD} & \textbf{Adam} & \textbf{SAM} & \textbf{\shortsamname} \\ \midrule
        $\mathcal{S}_{k\text{NN}}$ & 0.351 & 0.387 & 0.388 & 0.405 & \textbf{0.446} \\
        % Text acc.   & 0.796 & - & - & - & 0.800 \\ \hline
        % Image acc.  & 0.757 & - & - & - & 0.769 \\
        \bottomrule
    \end{tabular}
    \label{tab:alignment}
\end{table}
\begin{table}[t!]
    \centering
    \caption{Fine-tuned CLIP loss after convergence for varying learning rates with $\rho=0.01$. Remarkable divergence is shown in \textcolor{red}{red}.}
    \vspace{0.25cm}
    \begin{tabular}{l|ccc}
        \toprule
        & $1 \cdot 10^{-5}$ & $5 \cdot 10^{-5}$ & $1 \cdot 10^{-4}$ \\ \midrule
        \textbf{SAM} & 0.607 & \textcolor{red}{4.852} & \textcolor{red}{4.852} \\
        \textbf{\shortsamname} & \textbf{0.584} & \textbf{0.584} & \textbf{0.584} \\
        \bottomrule
    \end{tabular}
    \label{tab:alignment-fixed-perturbation}
\end{table}
Though SGD and Adam improves representational alignment with approx. 3-4\% compared to the pre-trained model - revealing the benefit of fine-tuning for improving downstream performance - seeking flatter minima with SAM and \shortsamname gives even better conceptual alignment, with \shortsamname finding a remarkably better solution than SAM. The fact that \shortsamname learns a conceptually better aligned representation space than SAM relates to \shortsamname's robustness to hyperparameter tuning; increasing the perturbation radius for \shortsamname generally enhances representational alignment, whereas SAM easily becomes unstable and take diverging steps, supposedly converging to a saddle point or local maxima. Thus, \shortsamname's conservativeness appears to also practically introduce robustness to hyperparameter tuning, potentially allowing for improved downstream performance.

\section{Related Work}

\paragraph{Extensions of SAM.} As mentioned, SAM contains an inherent scale-dependency problem \cite{dinh2017sharp}. A widely applied approach \cite{kwon2021asam} - namely, adaptive SAM (ASAM) - exploits a scale-invariant sharpness definition and modifies SAM by computing the worst-case perturbation in a normalized parameter space, typically by element-wise parameter scaling. Though improving on SAM on a variety of tasks including image classification, robustness to label noise and machine translation, the normalization operator is defined in an ad-hoc manner. This - along with the fact that the parameter space of deep neural networks is typically a statistical manifold that is not properly captured by Euclidean geometry - motivated the development of Fisher SAM \cite{kim2022fisher} that obtains slightly improved generalization performance on CIFAR-10 and CIFAR-100 than SAM and ASAM for a range of vision backbones. Recent advances include Riemannian SAM \cite{yun2024riemannian}; a general framework containing Fisher SAM that applies SAM on Riemannian manifolds by exploiting the tangent space and exponential maps. Applying the method requires predefining the manifold on which to optimize. They provide convergence analyses for SAM under the general formulation on Riemannian manifolds.

% SAM was justified by a PAC-Bayes generalization bound \cite{foret2020sharpness}, which has been criticized \cite{andriushchenko2022towards} as it relies on random perturbations rather than the worst-case perturbation used in practice. Another study of SAM \cite{behdin2023sharpness} argue that SAM can generalize better than GD due to its implicit bias of having lower bias.

\paragraph{Other approaches to find flat minima.} 
Stochastic weight averaging (SWA) \cite{izmailov2018averaging} exploits averaging of weights over iterations of SGD; SWA is shown to find wider solutions than SGD and empirically improved generalization compared to SGD for a variety of tasks. Meanwhile, SAM can find flatter optima than SWA in terms of eigenvalues of the Hessian \cite{kaddour2022flat}, yet, the better approach in terms of generalization performance is data- and task-specific with SWA e.g. working better for graph representation learning tasks than SAM. Other approaches include adjusting the gradient update by adding information from a random perturbation when the gradient norm is sufficiently low \cite{ahn2023escape} or applying SGD with dominant noise \cite{keskar2016large, jastrzkebski2017three, xing2018walk, zhu2018anisotropic, smith2020generalization, zhang2021understanding}, e.g. through large step sizes or small batches.

% \paragraph{Exploiting Geometry in Optimization.} 
% - NGD ?
% \vfill

\paragraph{The Monge metric.} 
% \citet{hartmann2022lagrangian} use the Monge metric but includes a geometry-controlling parameter, weighting the loss term in the definition of the manifold $\mathcal{M}$. \textcolor{red}{We implicitly fix this to 1 to avoid the need for tuning another hyperparameter.}
In recent years, the Monge metric has been used for various applications due to its simplistic form. It was comprehensively studied in context of geometric Markov Chain Monte Carlo (MCMC) sampling \cite{hartmann2022lagrangian} to replace the Fisher information matrix (FIM) for improving efficiency of geometric MCMC sampling. The study revealed that the metric has an inherent notion of curvature. A different study \cite{bergamin2024riemannian} proposed a Riemannian Laplace approximation for Bayesian neural networks (BNN) using the Monge metric for better capturing the underlying geometry of the intractable posterior. The method worked well on a variety of problems, yet the conservative nature of the metric implies suboptimal samples
%inexact samples that are biased towards the mode, originating from the conservativeness of the metric 
\cite{yu2023riemannian}. Instead \citet{yu2023riemannian} propose replacing the Monge metric with a metric based on the FIM, resulting in improved sample quality, which increases substantially the computational complexity.

\section{Conclusion}

% In this paper, we proposed \samname (\shortsamname), a novel reparameterization-invariant approach to sharpness-aware minimization. \shortsamname relies on the Monge metric; a Riemannian metric in the parameter space induced naturally by the training loss surface. The approach works under any modeling choice and only relies on a smoothness assumption of the loss function. Practically, we search for the worst-case adversarial perturbation according to the Monge metric, eventually resulting in perturbations being upper bounded by SAM's perturbation. This conservativeness makes \shortsamname behave like a mixture of SAM and GD; we saw implications of this in the context of deep neural networks. We discussed the benefits and pitfalls of conservatively exploring minima valleys rather than escaping and searching for globally flatter regions of the parameter space through experiments with fine-tuning using SAM and \shortsamname. While SAM easier escapes and might find better generalizing solutions, \shortsamname is more robust to the hyperparameter selection thus improving stability of the algorithm. Through an example considering improving multi-modal representational alignment by fine-tuning, SAM showed instability while \shortsamname found remarkably better solutions. This instability most likely arose due to SAM implicitly getting attracted to suboptimal equilibria like saddle points or maxima. We provide theoretical evidence of \shortsamname being less affected by this unwanted property.   

In this paper, we proposed \samname (\shortsamname), a novel reparameterization-invariant approach to sharpness-aware minimization. \shortsamname leverages the Monge metric, a Riemannian metric in the parameter space naturally induced by the training loss surface. The method works under any modeling choice, relying only on a smoothness assumption of the loss function.

Intuitively, the proposed \shortsamname searches for the worst-case adversarial perturbation according to the Monge metric, resulting in ``conservative'' perturbations bounded by SAM's perturbations. This makes \shortsamname behave as a mixture of SAM and GD, with the specific implications that we analyzed and observed empirically in the context of deep neural networks. 

The experiments revealed that while SAM escapes valleys more effectively to find flatter solutions with potentially better generalization, \shortsamname is more robust to hyperparameter selection, namely learning rate and pertubation radii, which improves the algorithmic stability. For fine-tuning tasks, particularly in improving multi-modal representational alignment, \shortsamname converged to solutions that improve the representational alignment remarkably while SAM exhibited instability, likely due to SAM's tendency to converge to suboptimal equilibria like saddle points or maxima. We further provided theoretical evidence supporting \shortsamname's resilience against getting attracted to suboptimal equilibria, compared to SAM.

\paragraph{Limitations and future work.} Though the conservativeness of \shortsamname guarantees properties like robustness and better resilience to saddle point attraction than SAM, it potentially prevents \shortsamname from escaping suboptimal valleys for finding globally flatter regions of the parameter space.

Similar to SAM, \shortsamname relies on a Taylor expansion to approximate adversarial perturbations, imposing ellipsoidal constraints rather than respecting the true geometric structure. Future work could investigate averaging the Monge metric over iterations or sampling it from the local neighborhood to better capture the local geometry.
% A more exact approach would be to search according to the geodesic ball, which is computationally expensive to obtain. An idea for future work is looking into solving this.
Another interesting direction is towards building a deeper understanding of the connection between the Monge and Fisher metrics, as previously hinted in related work \cite{yu2023riemannian}.

% \vfill
% \newpage

% % Acknowledgements should only appear in the accepted version.
% \section*{Acknowledgements}
% \textcolor{red}{Write the acknowledgements}

\vfill
\newpage
\section*{Impact Statement}
This paper introduces a novel optimizer designed to improve the generalization performance of machine learning models while considering the loss geometry. Our work advances optimization techniques in machine learning and has broad applicability. While the method requires more compute than standard gradient descent schemes, it is more robust to hyperparameter settings compared to SAM and might require less hyperparameter tuning when applied in practice. We do not identify any direct negative societal impacts of this contribution.

% ``This paper presents work whose goal is to advance the field of 
% Machine Learning. There are many potential societal consequences 
% of our work, none which we feel must be specifically highlighted here.''

\bibliography{monge-sam-arXiv}
\bibliographystyle{icml2025}

%%%%%%%%%%%%%%%%%%%%%%%%%%%%%%%%%%%%%%%%%%%%%%%%%%%%%%%%%%%%%%%%%%%%%%%%%%%%%%%
%%%%%%%%%%%%%%%%%%%%%%%%%%%%%%%%%%%%%%%%%%%%%%%%%%%%%%%%%%%%%%%%%%%%%%%%%%%%%%%
% APPENDIX
%%%%%%%%%%%%%%%%%%%%%%%%%%%%%%%%%%%%%%%%%%%%%%%%%%%%%%%%%%%%%%%%%%%%%%%%%%%%%%%
%%%%%%%%%%%%%%%%%%%%%%%%%%%%%%%%%%%%%%%%%%%%%%%%%%%%%%%%%%%%%%%%%%%%%%%%%%%%%%%
\newpage
\appendix
\onecolumn
\section{Proof of ODE describing local SAM dynamics}
\label{appendix:proof-ODE}

We extend the discrete SAM update to continuous time, hence establishing the SAM gradient flow:
\begin{equation*}
    \frac{d \boldsymbol{\theta}}{d t} = -f\left(z\left(\boldsymbol{\theta}\right)\right)
\end{equation*}
where $f\left(z\left(\boldsymbol{\theta}\right)\right) = \nabla \ell \left(z\left(\boldsymbol{\theta}\right)\right)$ and $ z\left(\boldsymbol{\theta}\right) = \boldsymbol{\theta} + \tilde{\rho} \ \nabla \ell \left(\boldsymbol{\theta}\right)$ and $\tilde{\rho}$ is the effective perturbation radius of either SAM or \shortsamname.
Next, we examine the dynamics in an $\boldsymbol{\epsilon}$-close neighborhood to $\boldsymbol{\theta}$ where $\boldsymbol{\epsilon}$ is a random vector drawn from $\mathcal{N}\left(\boldsymbol{0}, \sigma^2\mathbf{I}_d\right)$ with sufficiently small variance. We approach this by a Taylor expansion around $\boldsymbol{\theta}$:
\begin{align*}
    \frac{d \boldsymbol{\theta}}{d t} \approx \overset{\Ccancel[red]{\frac{d\boldsymbol{\theta}}{dt}} + \frac{d\boldsymbol{\epsilon}}{dt}}{\overbrace{\frac{d\left(\boldsymbol{\theta} + \boldsymbol{\epsilon}\right)}{d t}}} = - f\left(z\left(\boldsymbol{\theta} + \boldsymbol{\epsilon}\right)\right) &\approx -\biggl[f\left(z\left(\boldsymbol{\theta}\right)\right) + \nabla f\left(z\left(\boldsymbol{\theta}\right)\right)^\top \boldsymbol{\epsilon} \biggr] \\ \notag
    &= -f\left(z\left(\boldsymbol{\theta}\right)\right) - \nabla f\left(z\left(\boldsymbol{\theta}\right)\right)^\top \boldsymbol{\epsilon} \\ \notag
    &= \Ccancel[red]{\frac{d\boldsymbol{\theta}}{dt}} - \nabla f\left(z\left(\boldsymbol{\theta}\right)\right)^\top \boldsymbol{\epsilon}
\end{align*}
The dynamics at $\boldsymbol{\theta}$ cancel out and the $\boldsymbol{\epsilon}$-neighborhood is described by an ODE. Inserting $f$ and using the chain rule gives:
\begin{align*}
    \label{eq:stability-dynamics}
    % \frac{d\boldsymbol{\theta}}{dt} &\approx -\nabla_{\boldsymbol{v}} \left[ \nabla_{z\left(\boldsymbol{v}\right)} \ell \left(z\left(\boldsymbol{v}\right)\right)\right]^\top \boldsymbol{\varepsilon} \\ \notag
    \frac{d\boldsymbol{\epsilon}}{dt} &\approx -\nabla \left[ \nabla \ell \left(z\left(\boldsymbol{\theta}\right)\right)\right]^\top \boldsymbol{\epsilon} \\ \notag
    &= - \frac{\partial}{\partial \boldsymbol{\theta}} \left[ \nabla \ell \left(z\left(\boldsymbol{\theta}\right)\right)\right]^\top \boldsymbol{\epsilon} \\ \notag
    &= - \biggl(\frac{\partial}{\partial z\left(\boldsymbol{\theta}\right)} \nabla \ell\left(z\left(\boldsymbol{\theta}\right)\right)\cdot \frac{\partial z\left(\boldsymbol{\theta}\right)}{\partial \boldsymbol{\theta}} \biggr)^\top \boldsymbol{\epsilon} \\ \notag
    &= -   \biggl(\frac{\partial}{\partial z\left(\boldsymbol{\theta}\right)} \nabla \ell\left(z\left(\boldsymbol{\theta}\right)\right) \cdot \left[ \frac{\partial}{\partial \boldsymbol{\theta}}\boldsymbol{\theta} + \tilde{\rho} \frac{\partial}{\partial \boldsymbol{\theta}} \nabla \ell \left(\boldsymbol{\theta}\right) \right]\biggr)^\top \boldsymbol{\epsilon} \\ \notag
    % &= - \biggl(\nabla_{\boldsymbol{\theta}}^2 \ell\left(z\left(\boldsymbol{\theta}\right)\right) \cdot \left[ \frac{\partial}{\partial \boldsymbol{\theta}}\boldsymbol{\theta} + \tilde{\rho} \frac{\partial}{\partial \boldsymbol{\theta}} \nabla_{\boldsymbol{\theta}} \ell \left(\boldsymbol{\theta}\right) \right]\biggr)^\top \boldsymbol{\epsilon} \\ \notag
    &= - \ \underset{\mathbf{A}_{\rho}\left(\boldsymbol{\theta}\right)}{\underbrace{\biggl(\nabla^2 \ell\left(z\left(\boldsymbol{\theta}\right)\right) \cdot \bigl[ \mathbf{I}_d + \tilde{\rho} \nabla^2 \ell \left(\boldsymbol{\theta}\right) \bigr]\biggr)}}^\top \boldsymbol{\epsilon}
    % &= - \ \underset{\mathbf{A}_{\rho}\left(\boldsymbol{\theta}\right)}{\underbrace{ \biggl ( H_{\ell}\left(z\left(\boldsymbol{\theta}\right)\right) \cdot \bigl[ \mathbf{I}_d + \tilde{\rho} H_{\ell}\left(\boldsymbol{\theta}\right)\bigr] \biggr)^\top}} \ \boldsymbol{\epsilon}
\end{align*}
where $\nabla^2\ell\left(\boldsymbol{\theta}\right)$ is the Hessian at $\boldsymbol{\theta}$. Thus, SAM dynamics are governed by the local curvature at $\boldsymbol{\theta}$ along with the curvature at the perturbed parameter set $z\left(\boldsymbol{\theta}\right) = \boldsymbol{\theta} + \tilde{\rho} \ \lgrad$ when close to $\boldsymbol{\theta}$. Examining eigenvalues of $\mathbf{A}_{\rho}\left(\boldsymbol{\theta}\right)$ allows for addressing stability of SAM across the parameter space. Due to the minus sign, cases where the eigenvalues of  $\mathbf{A}_{\rho}\left(\boldsymbol{\theta}\right)$ are positive gives decaying dynamics, meaning $\boldsymbol{\epsilon} \rightarrow 0$, thus nearby points approach $\boldsymbol{\theta}$ and we call $\boldsymbol{\theta}$ an \emph{attractor} under SAM dynamics. \\

% \paragraph{Is SAM stable at equilibria?} 
% \subsubsection{Is SAM stable at equilibria?}
What \citet{kim2023stability} do, is to consider stability at equilibria points of the loss landscape, i.e. $\boldsymbol{\theta}^\ast$ where $\lgrad|_{\boldsymbol{\theta}=\boldsymbol{\theta}^\ast} = 0$. In such settings the SAM perturbation disappears as $z\left(\boldsymbol{\theta}\right)|_{\boldsymbol{\theta}=\boldsymbol{\theta}^\ast} = \boldsymbol{\theta}$ which results in
\begin{align*}
    \mathbf{A}_{\rho}\left(\boldsymbol{\theta}\right)|_{\boldsymbol{\theta}=\boldsymbol{\theta}^\ast} &= -\nabla^2\ell\left(\boldsymbol{\theta}^\ast\right) \cdot \bigl[\mathbf{I}_d + \tilde{\rho} \nabla^2 \ell\left(\boldsymbol{\theta}^\ast\right)\bigr] \\ \notag
    &= - \mathbf{Q} \mathbf{\Lambda} \mathbf{Q}^\top \cdot \bigl [\mathbf{I} + \tilde{\rho} \mathbf{Q} \mathbf{\Lambda} \mathbf{Q}^\top \bigr] \\ \notag
    &= - \mathbf{Q} \mathbf{\Lambda} \mathbf{Q}^\top + \tilde{\rho} \mathbf{Q} \mathbf{\Lambda} \mathbf{Q}^\top \mathbf{Q} \mathbf{\Lambda} \mathbf{Q}^\top \\ \notag
    &= - \mathbf{Q} \mathbf{\Lambda} \mathbf{Q}^\top + \tilde{\rho} \mathbf{Q} \mathbf{\Lambda}^2 \mathbf{Q}^\top \\ \notag
    &= - \mathbf{Q} \left[  \mathbf{\Lambda} + \tilde{\rho} \mathbf{\Lambda}^2 \right] \mathbf{Q}^\top    
\end{align*}
where the Hessian matrix, $\nabla^2\ell\left(\boldsymbol{\theta}^\ast\right)$, is real and symmetric and thus factorizes through the eigendecomposition to the form $H_{\ell}\left(\boldsymbol{\theta}^\ast\right)=\mathbf{Q} \mathbf{\Lambda} \mathbf{Q}^\top$, where $\mathbf{Q}$ is an orthonormal matrix, i.e. $\mathbf{Q}^T \mathbf{Q} = \mathbf{I}$ and $\mathbf{\Lambda} = \operatorname{diag}\left(\lambda_1, \dots \lambda_d\right)$ contains the eigenvalues. Specifically for equilibria, \citet{kim2023stability} determine a stability criteria by considering the eigenvalues of the Hessian. This means that all eigenvalues should satisfy:
$$
    -\left(\lambda_i + \tilde{\rho} \lambda_i^2\right) = - \lambda_i - \tilde{\rho} \lambda_i^2 < 0, \quad \forall \lambda_i \in \{\lambda_1, \dots, \lambda_d\}
$$
Since $\lambda_i^2 \geq 0$, isolating for $\tilde{\rho}$ gives:
\begin{equation*}
    \label{eq:stability-criteria-stationary}
    \tilde{\rho} > -\frac{1}{\lambda_i}, \quad \forall \lambda_i \in \{\lambda_1, \dots, \lambda_d\}
\end{equation*}

\vfill

\newpage
\section{Multi-modal representational alignment - full grid search}
\label{appendix:alignment}

\begin{table}[h!]
    \centering
    \caption{\textbf{CLIP performance after fine-tuning}. We evaluate pre-trained CLIP on the COCO Captions dataset as well as models fine-tuned to the Wiki dataset. We consider SGD, Adam, SAM and \shortsamname with varying $(\alpha, \rho)$-settings where $\alpha$ is the learning rate and $\rho$ is the perturbation radius. We present the loss and alignment scores at their optimal iterations, $T^\ast_{\text{loss}}$ and $T^\ast_{\text{alignment}}$, and mark the optimal loss values for each optimizer type in violet. We highlight diverging behavior as red cells with $^\ast$ denoting divergence behavior from the first fine-tuning step. Green cells represent more than $5\%$ improvement in terms of representational alignment, compared to the pre-trained CLIP model and the overall optimal performance is marked in bold text.}
    \vspace{0.25cm}
    \setlength{\tabcolsep}{11pt} % Adjust cell spacing
    \renewcommand{\arraystretch}{1.2} % Adjust row spacing
    \begin{tabular}{c|c|c|c|cc}
        \toprule \toprule
        \textbf{Optimizer} & \multicolumn{2}{c|}{\textbf{Params}} & \textcolor{black}{\textbf{Alignment} ($\uparrow$)} & \multicolumn{2}{c}{\textcolor{black}{\textbf{Loss} ($\downarrow$)}} \\ \cline{2-6}
        & $\alpha$          & $\rho$ & \textcolor{black}{$T=T^\ast_{\text{alignment}}$} & \textcolor{black}{$T=2000$} & \textcolor{black}{$T=T^\ast_{\text{loss}}$} \\ \midrule
        \textbf{None}           & -                 & -     & $^\ast$0.351 & - & $^\ast$0.699 \\ \hline \hline
        \multirow{2}{*}{\textbf{SGD}} 
                                & $1\cdot10^{-4}$   & -     & 0.376 & 0.596 & \cellcolor{violet!30!white} \textcolor{black}{0.591} \\ \cline{2-6}
                                & $1\cdot10^{-3}$   & -     & 0.387 & 0.663 & 0.634 \\ \hline \hline
        \multirow{2}{*}{\textbf{Adam}}   
                                & $1\cdot10^{-6}$   & -     & 0.376 & 0.631 & \cellcolor{violet!30!white} \textcolor{black}{0.595} \\ \cline{2-6}
                                & $5\cdot10^{-6}$   & -     & 0.388 & 0.600 & 0.600 \\ \hline \hline
        \multirow{15}{*}{\textbf{SAM}}  
                                & \multirow{5}{*}{$1\cdot10^{-5}$} & 0.005   & 0.397 & 0.597 & \cellcolor{violet!30!white} \textcolor{black}{0.584} \\ \cline{3-6}
                                &    & 0.01   & \cellcolor{teal!40!white} \textcolor{black}{0.405} & 0.623 & 0.607 \\ \cline{3-6}
                                &    & 0.02   & $^\ast$0.351 & \cellcolor{red!30!white} \textcolor{black}{4.849} & $^\ast$0.699 \\ \cline{3-6}
                                &    & 0.03   & $^\ast$0.351 & \cellcolor{red!30!white} \textcolor{black}{5.077} & 0.697 \\ \cline{3-6}
                                &    & 0.04   & $^\ast$0.351 & \cellcolor{red!30!white} \textcolor{black}{5.503} & $^\ast$0.699 \\
                                \cline{2-6}
                                & \multirow{5}{*}{$5\cdot10^{-5}$} & 0.005   & \cellcolor{teal!40!white} \textcolor{black}{0.401} & 0.593 & 0.591 \\ \cline{3-6} 
                                &    & 0.01   & $^\ast$0.351 & \cellcolor{red!30!white} \textcolor{black}{4.852} & $^\ast$0.699 \\ \cline{3-6}
                                &    & 0.02   & $^\ast$0.351 & \cellcolor{red!30!white} \textcolor{black}{4.852} & $^\ast$0.699 \\ \cline{3-6}
                                &    & 0.03   & $^\ast$0.351 & \cellcolor{red!30!white} \textcolor{black}{5.026} & $^\ast$0.699 \\ \cline{3-6}
                                &    & 0.04   & $^\ast$0.351 & \cellcolor{red!30!white} \textcolor{black}{4.882} & $^\ast$0.699 \\ 
                                \cline{2-6}
                                & \multirow{5}{*}{$1\cdot10^{-4}$} & 0.005   & \cellcolor{teal!40!white} \textcolor{black}{0.403} & 0.602 & 0.594 \\ \cline{3-6}
                                &    & 0.01   & $^\ast$0.351 & \cellcolor{red!30!white} \textcolor{black}{4.852} & $^\ast$0.699 \\ \cline{3-6}
                                & & 0.02   & $^\ast$0.351 & \cellcolor{red!30!white} \textcolor{black}{4.852} & $^\ast$0.699 \\ \cline{3-6}
                                &    & 0.03   & $^\ast$0.351 & \cellcolor{red!30!white} \textcolor{black}{4.870} & $^\ast$0.699 \\ \cline{3-6}
                                &    & 0.04   & $^\ast$0.351 & \cellcolor{red!30!white} \textcolor{black}{4.866} & $^\ast$0.699 \\ 
                                \hline \hline
        \multirow{12}{*}{\textbf{\shortsamname}}  
                                & \multirow{4}{*}{$1\cdot10^{-5}$} & 0.01   & 0.388 & 0.586 & 0.584 \\ \cline{3-6}
                                &    & 0.02   & \cellcolor{teal!40!white} \textcolor{black}{0.411} & 0.594 &  0.578 \\ \cline{3-6}
                                &    & 0.03   & \cellcolor{teal!40!white} \textcolor{black}{0.415} & 0.588 & \cellcolor{violet!30!white}{\textbf{0.576}} \\ \cline{3-6}
                                &    & 0.04   & \cellcolor{teal!40!white} \textcolor{black}{0.420} & 0.620 & 0.604 \\
                                \cline{2-6} 
                                & \multirow{4}{*}{$5\cdot10^{-5}$} & 0.01   &  \cellcolor{teal!40!white} \textcolor{black}{0.404} & 0.591 & 0.584 \\ \cline{3-6}
                                &    & 0.02   & \cellcolor{teal!40!white} \textcolor{black}{0.417} & 0.615 & 0.606 \\ \cline{3-6} 
                                &    & 0.03   & \cellcolor{teal!40!white} \textcolor{black}{0.433} & 0.602 & 0.587 \\ \cline{3-6}
                                &    & 0.04   & \cellcolor{teal!40!white} \textcolor{black}{\textbf{0.446}} & 0.620 & 0.615 \\ 
                                \cline{2-6}
                                & \multirow{4}{*}{$1\cdot10^{-4}$} & 0.01   & \cellcolor{teal!40!white} \textcolor{black}{0.405} & 0.602 & 0.584 \\ \cline{3-6}
                                & & 0.02   & \cellcolor{teal!40!white} \textcolor{black}{0.422} & 0.628 & 0.608 \\ \cline{3-6}
                                &    & 0.03   & $^\ast$0.351 & \cellcolor{red!30!white} \textcolor{black}{4.852} & $^\ast$0.699 \\ \cline{3-6}
                                &    & 0.04   & $^\ast$0.351 &  \cellcolor{red!30!white} \textcolor{black}{4.852} & $^\ast$0.699\\ 
                                \bottomrule \bottomrule
    \end{tabular}
    \label{table:performance-full}
\end{table}

%%%%%%%%%%%%%%%%%%%%%%%%%%%%%%%%%%%%%%%%%%%%%%%%%%%%%%%%%%%%%%%%%%%%%%%%%%%%%%%
%%%%%%%%%%%%%%%%%%%%%%%%%%%%%%%%%%%%%%%%%%%%%%%%%%%%%%%%%%%%%%%%%%%%%%%%%%%%%%%

\end{document}